# AN EVALUATION DATASET FOR LEGAL WORD EMBEDDING: A CASE STUDY ON CHINESE CODEX

Chun-Hsien Lin [1] and Pu-Jen Cheng [2]

[1] Department of Computer Science & Information Engineering, National Taiwan University, Taipei, Taiwan
`d03922030@csie.ntu.edu.tw; chunhsien.lin@gmail.com`

[2] Department of Computer Science & Information Engineering, National Taiwan University, Taipei, Taiwan
`pjcheng@csie.ntu.edu.tw`

## ABSTRACT

*Word embedding is a modern distributed word representations approach widely used in many natural language processing tasks. Converting the vocabulary in a legal document into a word embedding model facilitates subjecting legal documents to machine learning, deep learning, and other algorithms and subsequently performing the downstream tasks of natural language processing vis-à-vis, for instance, document classification, contract review, and machine translation. The most common and practical approach of accuracy evaluation with the word embedding model uses a benchmark set with linguistic rules or the relationship between words to perform analogy reasoning via algebraic calculation. This paper proposes establishing a 1,134 Legal Analogical Reasoning Questions Set (LARQS) from the 2,388 Chinese Codex corpus using five kinds of legal relations, which are then used to evaluate the accuracy of the Chinese word embedding model. Moreover, we discovered that legal relations might be ubiquitous in the word embedding model.*

## 1. INTRODUCTION

Word embedding is a modern distributed word representations approach and is widely used in many natural language processing tasks, such as semantic analysis [1], text classification [2], and machine translation [3, 4]. Most general-purpose applications mentioned above apply the evaluation benchmark dataset released by Google to assess their word embedding. For Chinese, the evaluation benchmark dataset was translated from the former by the Chinese Knowledge and Information Processing group (CKIP group) of the Academia Sinica Institute of Information Science. However, the general-purpose word embedding model cannot fully meet the task requirements of specific fields, such as biomedicine, financial opinion mining, the legal profession, etc. All have their own idiosyncratic terminology and thus should not be included in general-purpose word embedding models. In fact, in developing applications in these specific fields, collecting domain-relevant textual data to build a dataset and establish a word embedding model is often necessary. On the other hand, various languages also have their linguistic particularities. Sometimes, at the lexical word level, there will be no satisfactory translation. For example, plural nouns and verb tenses in English have no lexical equivalent in Chinese. Therefore, translating the existing general-purpose evaluation benchmark dataset directly as the target language word embedding evaluation benchmark has its limitations. As such, in addition to building a word embedding model for a specific field, it is very challenging to develop a corresponding evaluation benchmark dataset to evaluate the pros and cons of the word embedding model trained on a corpus of a specific field.

As the traditional Chinese text dataset lacked legal documents, we collected numerous articles containing legal provisions to serve as our dataset. Next, we established the basis of the legal relations dataset of this text dataset by conducting field expert review and induction. Finally, based on this dataset, we established an evaluation benchmark dataset for analogy reasoning. Using manual methods to create datasets is a last resort because of the serious drawbacks of higher costs, time expenditure, observer bias, and the potentially small size of the dataset, along with its innate irreproducibility. (We leave it to the future to develop an automatic method.) While many datasets are available for Chinese NLP experimentation, it is not easy to find legal domain experts to review and establish them. Given the absence of a Chinese legal data collection and evaluation benchmark dataset in our domestic environment, the manual method is our only recourse. While time and energy-consuming, the latter is nonetheless a pretty direct and intuitive approach. As regards the Chinese legal word embedding model and the evaluation benchmark dataset with legal relations, it will be very helpful in the future to introduce state-of-art technologies such as machine learning and deep learning into the Chinese legal profession application.

All the base data used in this experiment is collected from the 2,388 code articles found in the "Laws and Regulations Database of The Republic of China" (全國法規資料庫). Prior to the field expert manually reviewing the legal relationship between words in our vocabulary list, we use CKIPTagger, released by the CKIP group (aka CKIP Lab), to segment the sentences in the dataset as well as TF-IDF to sort and view the vocabulary. We also deploy gensim to combine the codex datasets mentioned above. Next, two modes, skip-gram and CBoW, respectively, are used to generate word embedding models. Then, TensorBoard is used to visually assist manual inspection of the relevance of each word in the vector space. Finally, we manually select the appropriate legal relationship vocabulary to build the legal analogy reasoning questions (Legal Analogical Reasoning Questions Set, LARQS) in order to evaluate the accuracy of the word embedding model trained from the codex corpus.

Various evaluation methods have been proposed to assess the qualities of word embedding models [5]. However, there is still no scientific consensus on which evaluation method should be used for word embedding models [6]. Therefore, this paper applied the simple algebraic calculation method as analogical reasoning proposed by Mikolov [7] to evaluate the accuracy of word embedding models. Applying Mikolov's simple algebraic calculation method is intuitive, but also requires an evaluation benchmark dataset with broader relationship coverage for the accuracy of the evaluation to be more objectively determined. As such, this paper attempts to establish an 1,134 Legal Analogical Reasoning Questions Set (LARQS) with five categories of legal relationships from the corpus collecting 2,388 Chinese legal codices so as to evaluate the accuracy of Chinese word embedding from the perspective of legal relations.

Our main contributions in this work can be summarized as follows:

- We collected the legal provisions of 2,388 laws and regulations promulgated and implemented in Taiwan as a codex corpus for experimentation. Next, we trained a legal word embedding model based on this corpus and released it to the public for future research.

- We established and released a 1,134 Legal Analogical Reasoning Questions Set (LARQS) with five categories of legal relationships from the corpus mentioned above to evaluate the accuracy of Chinese word embedding from the perspective of legal relations.

- We assessed the accuracy of several word embedding models with the Google evaluation benchmark dataset. Our conclusion is that, generally speaking, the evaluation benchmark dataset is not suitable for the legal profession.

- We discovered that legal relations may be ubiquitous in the word embedding model.

This paper is organized as follows: Section 2 provides an essential background for Chinese word segmentation, which is the first and most critical step in Chinese natural language processing, and reviews the development and research of word embedding models in Chinese. Moreover, we review several benchmark datasets for the word embedding model and the evaluation approach used in this paper. In section 3, we give a brief description of the dataset source of the experiment, the established procedures, and the tools used in this paper. Section 4 explains the meanings implied by legal terms and their interrelationship on the conceptual level. Section 5 describes the evaluation dataset proposed in this paper and the experimental results of other evaluation datasets on different Chinese word embedding models. Finally, section 6 discusses the possibility of working with the LARQAS benchmark dataset in future

## 2. RELATED WORKS

The most famous approach to the distributed representation of words in a vector space is the word embedding model proposed by Mikolov [8]. Capable of quickly producing compact vectors, this algorithm is widely used by the natural language processing community. On the other hand, when it comes to the distributed representation of words, simple algebraic calculations can be used to obtain the offset value of each word pair. This approach can achieve the effect of acquiring "linguistic regularities" by analogical reasoning, as in "king - man + woman ≈ queen". Given the simplicity of this calculation, some scholars have also tried to capture the semantics of biomedical concepts [9], and extract mentions of adverse drug reactions from informal text found on social media [10]. In addition to obtaining semantic rules, we can also use the vector calculation method described above to induce morphological transformations between words [11]. Moreover, Ash [44] has similarly shown that the word embedding model trained on a corpus of statutes result in vector("corporate income tax") – vector("corporation") + vector("person") ≈ vector("personal income tax"). Based on the premise that these words' "linguistic regularities" are assumed to have linear relations, vector offset is used to capture the syntactic and semantic rules of word analogy reasoning. However, the semantic coverage of this approach is limited [12, 13]. This is hardly surprising because even Google's public evaluation benchmark dataset contains only 14 linguistic categories.

Although there has been considerable research into the evaluation of Chinese word embedding, most of it has been based on the aforementioned algebraic calculation as analogy reasoning of the word embedding vector offset value [14], which facilitates exploring the morphology or semantics of Chinese words. Research into the concepts and correlation between legal terms and word embedding models has been comparatively rare. Factors contributing to this include the difficulty of obtaining datasets and appropriate field experts to participate in such research.

When it comes to Chinese natural language processing workflow, the first obstacle is Chinese word segmentation [15]. Moreover, the accuracy of Chinese word segmentation results has varying degrees of impact on subsequent downstream Chinese natural language processing tasks [16]. To reduce the error rate in these tasks, it is thus imperative to choose the most appropriate Chinese word segmentation tools. In this regard, many researchers have already produced numerous outstanding study findings in this field [17]. As such, for this paper we were able to select the most appropriate available tools to handle Chinese word segmentation.

To assess and compare the performance of the benchmark set, we build two sets of word embedding models, one with codex articles, the other with the Chinese version of Wikipedia. After establishing the word embedding model, the next step is to explore means for evaluating its accuracy. Subsequently, we assess the accuracy of candidate word embedding models using the Google evaluation benchmark dataset translated by the CKIP Group of the Academia Sinica Institute of Information Science, the CA8 evaluation benchmark dataset released by Shen Li [14], and the legal analogy dataset proposed in this paper (Legal Analogical Reasoning Questions Set, LARQS). Using these evaluation benchmark datasets, we then apply the simple algebraic calculation analogical reasoning approach proposed by Mikolov [7] and compare it with these benchmarks and word embedding models.

### 2.1. Chinese Word Segmentation

Chinese word segmentation is the first and most critical step in Chinese natural language processing. Many researchers have proposed methods to solve this necessary process [18-20]. Moreover, a deep neural network has recently been introduced to work on this issue and achieved conspicuously superior results [21]. There are many Chinese word segmentation tools. Python is the primary programming language for this paper's experiment. In the early stage of the experiment, we used the word segmentation tool "Jieba", a famous Chinese word segment package for Python whose primary advantage is speed. However, Jieba's development was based on simplified Chinese. When processing large numbers of traditional Chinese documents, it turned out to be outperformed by CKIPTagger (released by the Chinese Knowledge Information Processing Group of the Academia Sinica Institute of Information Science). With our objective being to handle Chinese codex articles, and the words in the codex belonging to the legal professional field, Jieba's word segmentation results proved even more unsatisfactory. After manually reviewing the segment results of the word segmentation tools Jieba and CKIPTagger, we finally chose CKIPTagger as the word segmentation tool for this experiment because of its superior accuracy with traditional Chinese word segments. Figure 1 is an example of word segmentation in a Chinese law article.

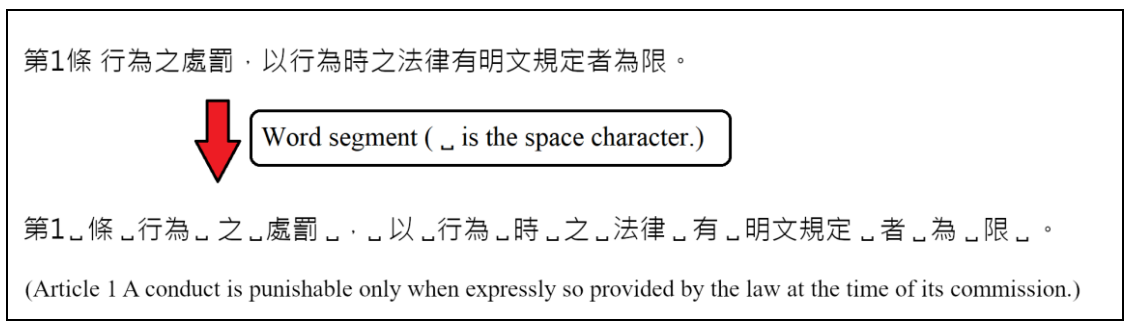

Figure 1. Word segment a Chinese law article with a space character.

### 2.2. The Model Architectures for Word Embedding

In many Natural Language Processing tasks, the representation of a lexical word into a numerical form that the computer can calculate is a fundamental and necessary pre-processing procedure. The conventional one-hot vector is an intuitive approach, but the distributed word representations can represent lexical words with a low-dimensional dense vector and embed more latent information [22]. Distributional word representations based on co-occurrence information between lexical words are LSA [23] and LDA [24]. There exist various approaches for obtaining the distributed dense real-valued vector from unlabelled text corpora. The most famous of which are Glove [25] and word2vec Skip-gram [7]. Recent work has shown that transforming lexical words into a distributed dense real-valued vector in a geometric space can capture the semantic "meaning" of a single word embedded in the vector via simple algebraic calculations and other methods. [26-29].

The distributed word embedding architectures proposed by Mikolov are Skip-gram and CBoW. Due to its success in modelling English documents, word embedding has been applied to Chinese text. Benefiting from the internal structural information of Chinese characters, many studies tried to enhance the quality of Chinese word embeddings with radicals [30-32], sub-word components [33, 34], glyph features [35], strokes [36], and pronunciation [37]. To limit the scope of this paper, we choose Skip-gram because, after comparing the word embedding model established by the two corpora used in this experiment, we found Skip-gram to have the best performance on average.

### 2.3. Existing public benchmark datasets and finding similar words

As regards evaluating word embedding models, Google has released a test set consisting of about 20,000 questions-words plus their syntactic and semantic relations. However, it only offers morphological and semantic relations for both analogical reasoning and capturing linguistic regularities, but does not explore other relationships in great breadth. For example, as mentioned above, the semantic relations of biomedical concepts or legal term relations were not included in the Google test set. The test set has a total of 14 relationships, encompassing 9 morphological and 5 semantic categories, and a total of 19,544 questions. The Chinese Knowledge Information Processing Group of the Academia Sinica Institute of Information Science (aka CKIP Group) also translated the Google test set [38], with 11,126 questions. A summary of the content released by Google and translated by the CKIP Group is shown in Table 1:

Table 1. A summary of the benchmark released by Google and translated by the CKIP Group

| Relation | Example | Translated to Chinese |
| --- | --- | --- |
| capital-common-countries | Beijing China<br>Tokyo Japan | 北京 中國<br>東京 日本 |
| capital-world | Ankara Turkey<br>Cairo Egypt | 安卡拉 土耳其<br>開羅 埃及 |
| Currency | USA Dollar<br>Europe Euro | 美國 美元<br>歐洲 歐元 |
| city-in-state | Chicago Illinois<br>Honolulu Hawaii | 芝加哥 伊利諾州<br>檀香山 夏威夷州 |
| family | brother sister<br>king queen | 兄弟 姐妹<br>國王 皇后 |
| gram1-adjective-to-adverb | amazing amazingly | (No Chinese mapping) |
| gram2-opposite | decided undecided<br>efficient inefficient | 決定 未決<br>有效率 低效率 |
| gram3-comparative | bad worse | (No Chinese mapping) |
| gram4-superlative | bad worst | (No Chinese mapping) |
| gram5-present-participle | code coding | (No Chinese mapping) |
| gram6-nationality-adjective | Albania Albanian<br>China Chinese | 阿爾巴尼亞 阿爾巴尼亞人<br>中國 中國人 |
| gram7-past-tense | dancing danced | (No Chinese mapping) |
| gram8-plural | banana bananas | (No Chinese mapping) |
| gram9-plural-verbs | decrease decreases | (No Chinese mapping) |

Leveraging word analogical reasoning as an evaluation benchmark is fascinating and has potential as an approach for discovering linguistic relations [39]. In 2010, Turney and Pantel proposed an extensive survey of tasks that could be considered as a reference for measuring the performance of word embedding [40]. In 2013, Mikolov showed that proportional analogies (a is to b as c is to d) could be solved by finding the vector closest to the hypothetical vector calculated as c - a + b (i.e., king - man + woman ≈ queen). The assumption is that a "well-trained" word embedding encodes linguistic relations so that they are identifiable via linear vector offset [7, 41]. Although various evaluation methods are mentioned in the literature above, however, to limit the scope of this paper, we choose the original approach proposed by Mikolov to find out the hidden vector d. This simple calculation is quite helpful for subsequent manually finding similar words to build the LARQS benchmark dataset.

## 3. DATASET CONSTRUCTION AND TOOLS USED IN THE EXPERIMENT

The dataset (Chinese codex dataset) established in this experiment is collected from the Chinese provisions of domestic laws and regulations published in the "Laws and Regulations Database of The Republic of China" (全國法規資料庫). We collected 2,388 statutes and regulations, consisting of 73,365 articles (excluding the article section names, the total is 67,727 articles). Next, word segmentation with CKIPTagger and terms with a word frequency of less than five are manually checked. Ultimately, a dataset emerges consisting of 5,830,894 words. Additionally, using the dataset included in the Chinese version of Wikipedia up to December 1, 2019 and the same parameters as in the following, we established 19 different dimensional word embedding models for comparison.

The tool used to build the Chinese word embedding models in this paper is gensim. Even using this existing tool, the hyper parameters for the training word embedding model greatly influence the final quality [42, 43]. Therefore, to explore the best possible hyper parameters, we selected several hyper parameters such as the vector size of word embedding, iterations, and the window sizes for the experiment. We set the following parameters to train it: vector size from 100 to 1000 dimensions, with an interval of every 50 dimensions, and the number of iterations being 1, 10, 100, 200, and 1000. Chiefly using the skip-gram architecture, and with the word window size being 7 and the smallest token size set as a single Chinese character, we built 19 Chinese word embedding models using these parameters from the dataset mentioned above.

Figure 2 is a schematic illustration of the construction process for the legal word embedding modelling and the evaluation benchmark dataset. The process can be divided into three major steps. The first is to collect the textual data of the codex articles, after completing pre-processing such as data cleaning and word segmentation. Then the data is converted into a dataset whose format can be used by subsequent tools. Next, we set various hyper parameters to adjust the word embedding training tool and then import the dataset into the tool for training. The third step is having legal experts provide various legal relationships and corresponding vocabulary references from codices, using TensorBoard to assist the legal experts in reviewing the vocabulary in the word embedding models, and, finally, establishing the LARQS evaluation dataset manually. When using TensorBoard and identifying Chinese lexical words inappropriate due to word segmentation, we use the dotted line in Figure 2 to return to the pre-processing step to manually fine-grain the word segmentation, and then go back to the second step to re-train the word embedding model.

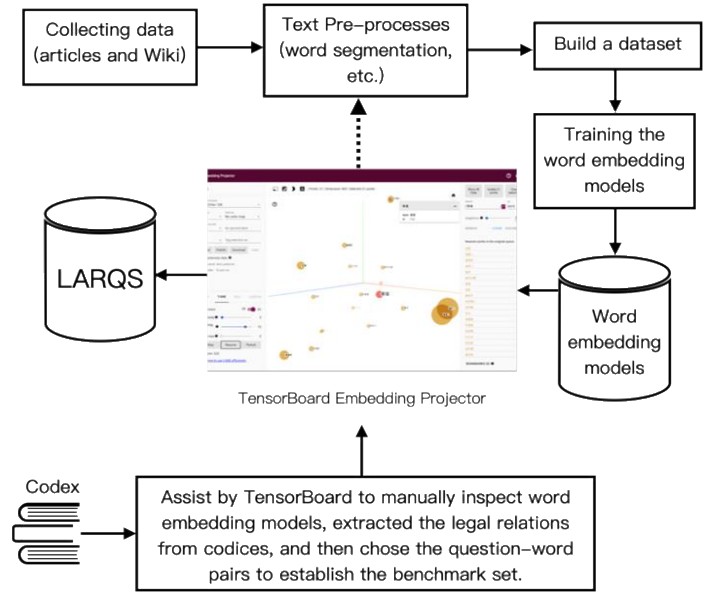

Figure 2. The pipeline of word embedding modeling and benchmark dataset construction.

## 4. CATEGORIES OF LEGAL TERM RELATIONS

Through expert manual inspection, we extracted five categories of legal relations from 2,388 codices to experiment with analogical reasoning. We then established a dataset consisting of 1,134 legal analogical reasoning questions-words (Legal Analogical Reasoning Question Set, LARQS). The five categories of legal term relations included "Prime and Deputy", "Rights and obligations", "Execute and staff", "Operation", and "Rights". Table 2 is a sample of this.

Table 2. A sample of Legal Analogical Reasoning Question Set

| Relation Category | Example | Questions | Words |
|---|---|---|---|
| Prime and Deputy (正副關係) | 正首長(chief),副首長(deputy chief) 總統(president),副總統 (vice president) | 870 | 30 |
| Rights and obligations (權利義務相對人) | 債權人(creditor),債務人(debtor) 典權人 (dien-holder),出典人(dian-maker) | 42 | 7 |
| Execute and staff (執業行為與人員) | 查核(audit),會計師(certified public accountant) 競選(campaign,),候選人(candidate) | 42 | 7 |
| Operation (客體操作行為) | 疫苗(vaccine),接種 (vaccination) 森林(forest),墾殖(cultivating) | 90 | 10 |
| Rights (權利與權利人) | 債權(claims),債權人(creditor) 電路布局權(circuit layout rights),電路布局權人(circuit layout right owner) | 90 | 10 |

After estimating the accuracy of these two embedding models (the legal word embedding model and the Chinese Wikipedia word embedding model) with LARQS, we found that the various legal relations listed in Table 2 are common in the legal word embedding model and that legal relations exist in the general word embedding model too. This finding is presented in this paper and will be discussed in the next section.

# 5. EXPERIMENTAL RESULTS

We use the accuracy function by gensim provided and various benchmarks sets in this paper. The accuracy of the codex word embedding model, established in skip-gram, and the experimental results from evaluating with the LARQS benchmark dataset are exhibited in Figure 3. In this same figure, the size of the word embedding affects accuracy, while iteration times are an additionally important parameter. In this experiment, the number of iterations range from 100 to 200, the word embedding size ranges from 700 to 850 dimensions, and the average performance accuracy is as high as 65%. However, when the iterations were set to 1000, the accuracy proved inferior to the iterations being set to 100.

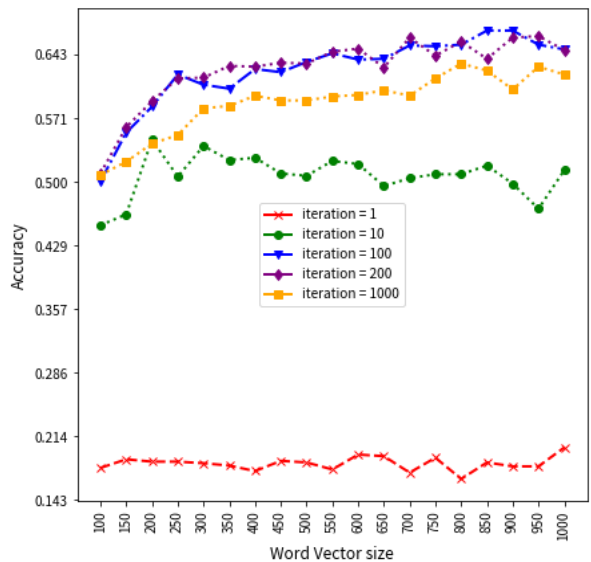

Figure 3. Accuracy affected by word embedding size and training iterations (skip-gram)

The word embedding model proposed by Mikolov has another architecture, CBoW. This paper uses the CBoW architecture to establish a word embedding model for the same codex dataset and then uses this paper's LARQS evaluation benchmark dataset to evaluate accuracy. The average accuracy is not superior to that of Skip-gram architecture, as shown in Figure 4. Therefore, the subsequent experiments in this paper use the Skip-gram architecture word embedding model to evaluate accuracy.

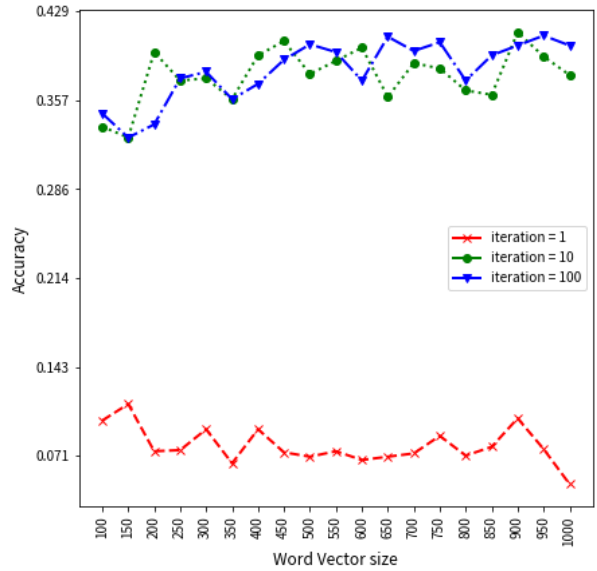

Figure 4. Accuracy affected by word embedding size and training iterations (CBoW)

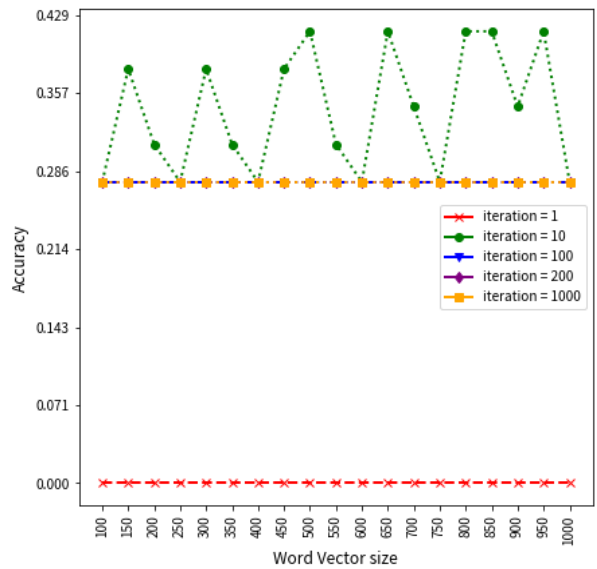

Figure 5. Accuracy affected by word embedding size and training iterations (zh-tw Google set)

To observe the performance differential between the LARQS and other public benchmark datasets, we first selected the evaluation benchmark dataset released by Google and translated by CKIP Group. As can be seen from the experimental results in Figure 5, when the Google evaluation benchmark dataset translated by CKIP Group is subjected to different training iterations, the accuracy of the Chinese Codex word embedding model is not outstanding.

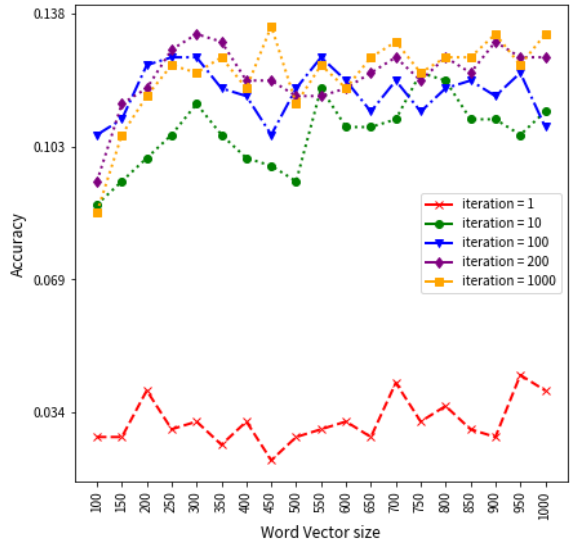

Figure 6. Accuracy affected by word embedding size and training iterations (CA8/morphological)

Figures 6 and 7 present our evaluation of our Chinese codex word embedding model with the CA8 evaluation benchmark datasets published by Shen Li [14]. The CA8 evaluation benchmark datasets fall into two categories: morphological and semantic. Figure 6 shows the experimental results of evaluating the accuracy of the Chinese Codex word embedding model with the CA8 morphological evaluation benchmark dataset. The performance of the Chinese codex word embedding model is very low. Figure 7 is the CA8 semantic benchmark dataset, which is used to evaluate the accuracy of the Chinese Codex word embedding model and whose performance is similarly less than ideal.

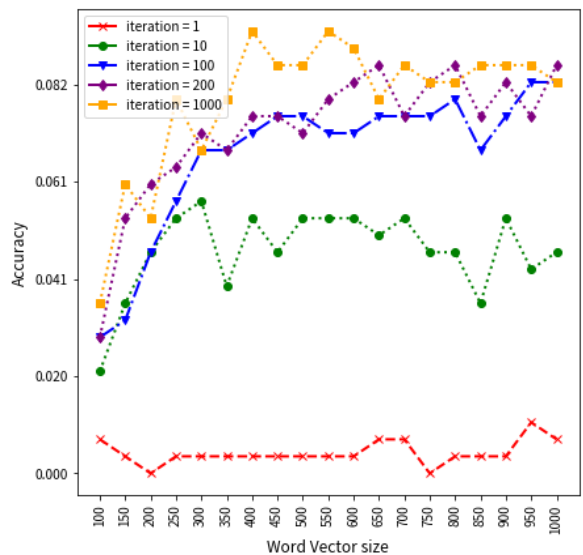

Figure 7. Accuracy affected by word embedding size and training iterations (CA8/semantic)

Figure 8 summarizes the experimental results of evaluating the aforementioned datasets. Each of the previous benchmark dataset experiments using the word embedding model established by 2,388 codices is collected in this paper. The size of the word embedding models ranged from 100 to 1000, with iterations numbering 100. Next, the simple algebraic calculation method that Mikolov [7] proposed to evaluate accuracy is applied. As Figure 8 reveals, the top accuracy achieved with LARQS in this paper is 67.02% using word embedding in 850 dimensions. This significantly outperforms the linguistic-based CA8 evaluation benchmark dataset by 8.2%, and also improves upon Google's evaluation benchmark dataset translated by the CKIP Group by some 27.58%. Table 3 illustrates the accuracy achieved with different word embedding sizes in the same Chinese Codex word embedding model with regard to each evaluation dataset.

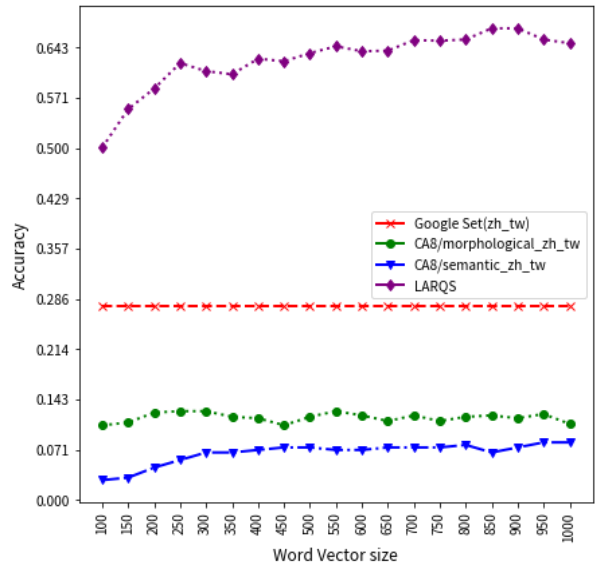

Figure 8. Estimating accuracy of codex word embedding model with different benchmark

Table 3. The accuracy achieved with different word embedding sizes in the Chinese Codex word embedding model with regard to each benchmark dataset.

| Benchmark Dimension | LARQS | Google Set | CA8/morphological | CA8/semantic |
|---|---|---|---|---|
| 100 | 0.5 | **0.2758** | 0.1064 | 0.0285 |
| 200 | 0.5846 | 0.2758 | 0.1244 | 0.0464 |
| 300 | 0.6093 | 0.2758 | **0.1265** | 0.0678 |
| 400 | 0.6269 | 0.2758 | 0.1164 | 0.0714 |
| 500 | 0.6349 | 0.2758 | 0.1184 | 0.075 |
| 600 | 0.6375 | 0.2758 | 0.1204 | 0.0714 |
| 700 | 0.6534 | 0.2758 | 0.1204 | 0.075 |
| 800 | 0.6543 | 0.2758 | 0.1184 | **0.0785** |
| 850 | **0.6702** | 0.2758 | 0.1205 | 0.0679 |
| 900 | 0.6702 | 0.2758 | 0.1164 | 0.075 |
| 1000 | 0.649 | 0.2758 | 0.1084 | 0.0821 |

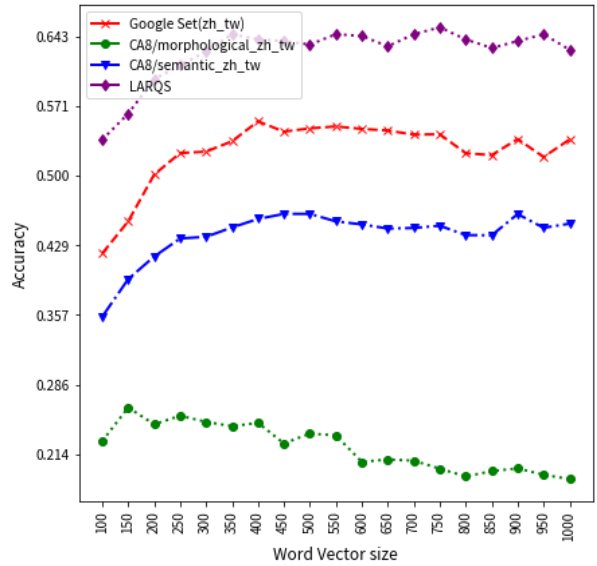

Figure 9. Estimating accuracy of Wikipedia (zh-tw) word embedding model with different benchmark

To explore the legal relationship between the words in the word embedding model, we also collected the Chinese version of Wikipedia, which covers a wide range of content, as a dataset. This experiment builds 19 differently-sized word embedding models based on the Chinese version of Wikipedia up to December 1, 2019. The word embedding size ranges from 100 to 1000, with 50 dimensions as the interval range for each model. Next, applying the LARQS in this paper, the Google benchmark set translated by Academia Sinica and the two CA8 benchmark sets (translated from simplified to traditional Chinese) are used to estimate the accuracy of the Wikipedia word embedding model. Figure 9 shows the accuracy of the Wikipedia word embedding model evaluated using different benchmark sets. In the best-case scenario for LARQS, i.e., when the word embedding size is 750, an accuracy level of 65.24% can be achieved. This is higher than the translated Google evaluation data set whose peak accuracy is 55.59% when the word embedding size is 400. Table 4 illustrates the accuracy of various word embedding sizes of the Wikipedia Chinese word embedding model, estimated using different benchmark datasets. Reviewing the results of this experiment (even for general documents, and estimating the word embedding model established with the LARSQ dataset for legal relations), when compared with the 14 categories in the benchmark dataset released by Google (i.e., in linguistics, capital-common-countries, currency-names-and-countries, and using the CA8 benchmark dataset based purely on linguistics), the LARQS dataset in this paper can better demonstrate the universality of legal relations between vocabulary relations in the word embedding model.

Table 4. The accuracy of various word embedding sizes of the Wikipedia Chinese word embedding model, estimated using different benchmark datasets.

| Benchmark Dimension | LARQS | Google Set | CA8/morphological | CA8/semantic |
|---|---|---|---|---|
| 100 | 0.5372 | 0.4204 | 0.2278 | 0.3554 |
| 200 | 0.5993 | 0.5013 | 0.2450 | 0.4172 |
| 300 | 0.6277 | 0.5249 | 0.2474 | 0.4374 |
| 400 | 0.6401 | 0.5559 | 0.2464 | 0.4557 |
| 500 | 0.6348 | 0.5488 | 0.2355 | 0.4609 |
| 600 | 0.6436 | 0.5482 | 0.2063 | 0.4501 |
| 700 | 0.6454 | 0.5423 | 0.2076 | 0.4465 |
| 750 | 0.6525 | 0.5427 | 0.1994 | 0.4488 |
| 800 | 0.6401 | 0.5233 | 0.1914 | 0.4392 |
| 900 | 0.6383 | 0.5378 | 0.1996 | 0.4608 |
| 1000 | 0.6294 | 0.5373 | 0.1886 | 0.4509 |

## 6. FUTURE WORK AND CONCLUSION

Applying simple algebraic calculations to obtain the deviation values between word vectors, combined with manually selected Chinese words (vocabulary), and mining the relationship between the vocabulary in the word embedding model with its legal relationship, is a paradigm of especial interest proposed in this paper. Our experiment's results indicate that legal relations might be ubiquitous in the word embedding model based on legal provisions, aka codices, and word embedding models covering a wider range of content. We are also publishing the LARQS benchmark dataset and the Chinese codex word embedding model from our experiment to expedite carrying out NLP tasks related to Chinese law. In the future, we hope that automated methods will be developed to unearth additional legal relationships and enrich the LARQS dataset proposed in this paper. These, applied with this experiment's word embedding model, combined with more advanced machine learning algorithms, could be applied to excellent effect on more complex NLP tasks such as document generation, automatic contract review, document classification, and machine translation.

**Authors**

Chun-Hsien Lin is currently a Prosecutorial Affairs Officer of Taiwan High Prosecutors Office and pursuing the Ph.D. degree from the National Taiwan University of the Department of Computer Science & Information Engineering. His research is mainly on AI and law, natural language processing, machine learning, and so on.

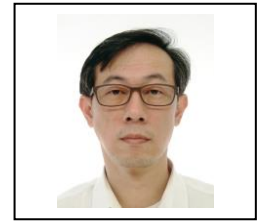

Pu-Jen Cheng received his Ph.D. in Computer Science at National Chiao Tung University in 2001. He went to the Institute of Information Science, Academia Sinica, as a Postdoctoral Fellow for more than four years. Starting from August 2006, he joined the department of Computer Science and Information Engineering faculty at the National Taiwan University, and also jointly appointed at the Graduate Institute of Networking and Multimedia, National Taiwan University. He is a member of the ROC Phi Tau Phi Scholastic Honor Society and the ACM/SIGIR.

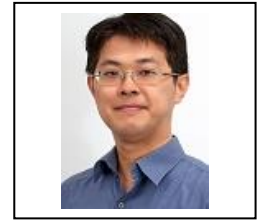